\documentclass[runningheads]{llncs}

\usepackage{accv}

\usepackage{accvabbrv}

\usepackage{microtype}
\usepackage{graphicx}
\usepackage{subcaption}
\usepackage{booktabs} % for professional tables
\usepackage{hyperref}
\usepackage{multirow}
\usepackage{bbm}
\usepackage{dsfont}
\usepackage{algorithm}
\usepackage{algorithmic}
\usepackage{amsmath}
\usepackage{amssymb}
\usepackage{mathtools}
\usepackage{booktabs} % For professional lines (\toprule, \midrule, \bottomrule)
\usepackage{makecell} % For breaking long headers into multiple lines
\usepackage{amsmath}  % For clean math formatting
\usepackage[accsupp]{axessibility}  % Improves PDF readability for those with disabilities.
\usepackage[capitalize,noabbrev]{cleveref}
\usepackage[textsize=tiny]{todonotes}

\usepackage{hyperref}

\usepackage{orcidlink}

\begin{document}

\title{PathoS2VD: From 2D Histopathology to Pseudo Z-Stack Representations} 

\titlerunning{PathoS2VD}

% Second Author\inst{2,3}\orcidlink{1111-2222-3333-4444} \and
% Third Author\inst{3}\orcidlink{2222--3333-4444-5555}}

\author{Jiamu Wang\inst{1} \and
Keunho Byeon\inst{1} \and 
Ji-Uk Kim\inst{1} \and 
Jin Tae Kwak\inst{1} \and 
Heounjeong Go\inst{2} 
}

% index{Wang, Jiamu}
% index{Byeon, Keunho}
% index{Kwak, Jin Tae}

\authorrunning{J. Wang et al.}

% Springer Heidelberg, Tiergartenstr.~17, 69121 Heidelberg, Germany

\institute{
School of Electrical Engineering, Korea University, Seoul 02841, Korea\\
\email{\{taurusmumu, bkh5922, kluki, jkwak\}@korea.ac.kr}
\and
Department of Pathology, Asan Medical Center, University of Ulsan College of Medicine, Seoul 05505, Korea\\
\email{damul37@amc.seoul.kr}
}

\maketitle

\begin{abstract}
  For decades, diagnostic pathology has relied on thin two-dimensional (2D) tissue sections, despite biological tissues exhibiting inherently three-dimensional (3D) morphological organization. Because a single section captures only one focal plane, important morphological cues along the z-axis are inevitably discarded, limiting tissue representation for both human interpretation and computational analysis. In this work, we propose a generative framework that learns z-axis morphological priors from pathology z-stack data, offering a cost-effective alternative to dedicated volumetric imaging, and transfers this knowledge to conventional single-plane histopathology. Specifically, an image-to-video (I2V) latent diffusion model (LDM) is trained to synthesize a pseudo z-stack from a single 2D image, capturing plausible morphological variations across adjacent focal planes rather than reconstructing physical tissue volumes. To enable deployment on large-scale clinical datasets without native z-stack acquisition, we further introduce a prior-preserving mixture fine-tuning strategy that adapts the model to new domains while retaining the learned z-axis morphological prior. Extensive experiments demonstrate that the synthesized pseudo z-stacks exhibit realistic cross-plane morphological continuity and consistently improve downstream prostate cancer classification compared with conventional 2D representations. Our results suggest that learning transferable z-axis priors from limited z-stack data provides a practical way to enrich routine pathology images with complementary morphological information. Code: https://github.com/QuIIL/PathoS2VD.
\end{abstract}

\section{Introduction}
\label{sec:intro}

\begin{figure*}[t]
    \centering
    \includegraphics[width=1\textwidth]{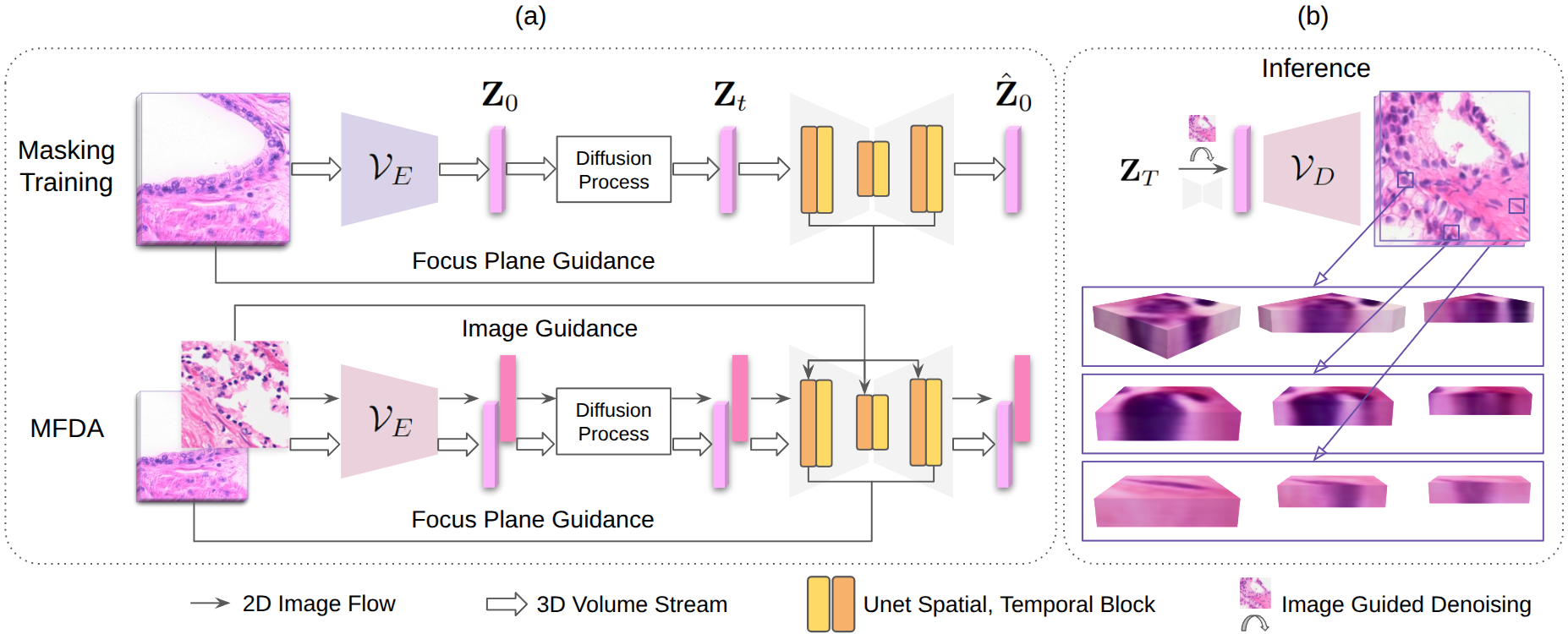}
    \caption{We aim to synthesize plausible tissue volume along the z-axis surrounding a given image input (right). We first train an I2V diffusion model with the masked z-stack dataset to capture the z-axis prior in a sharp range (left top. Next, we finetune the generative model with the z-stack and the cancer image dataset jointly to adapt the z-axis prior onto the target image domain (left bottom).}
    \label{fig:fig1}
\end{figure*}

Histopathology examination of biopsies and surgical specimens underpins the definitive diagnosis for many diseases. Despite major advances in tissue processing as well as digital and computational pathology, routine clinical practice still relies on thin two-dimensional (2D) tissue sections as the primary substrate for interpretation. As only a small fraction of a three-dimensional (3D) specimen is observed, this workflow is inherently undersampled and can miss morphological information along the tissue depth, potentially leading to biased or uncertain diagnoses \cite{harnessing_nondestructive}. Collapsing complex 3D tissue structures into a 2D single plane can obscure diagnostically relevant morphology and spatial heterogeneity, limiting the reliability of both human assessment and computational models trained on single-plane data.

Recent advances in non-destructive 3D Imaging, such as open-top light-sheet (OTLS) microscopy \cite{otls_paper} and micro-computed tomography (microCT) \cite{microct_paper}, offer volumetric acquisition and visualization of intact tissues with rich structural context. Such modalities provide complementary morphological information across multiple focal planes that is difficult to infer from a single 2D section. For example, in a 2D single plane, poorly formed glands may appear ambiguous in prostate pathology \cite{engineering_future} and infiltrative tumor extensions that may appear only as isolated buds \cite{cancer_invasion_3d}. Deep Learning (DL) approaches have therefore emerged to leverage this volumetric information effectively. Recent pipelines like TriPath \cite{tripath} utilize 3D feature extractors with attention-based aggregation for patient-level predictions, demonstrating the superior ability of 3D features over 2D analysis. Similarly, CARP3D \cite{carp3d} proposes a computational triage strategy that efficiently sifts through volumetric data using 2D feature extractors and multiple instance learning (MIL) \cite{mil} to identify high-risk planes.

However, the clinical adoption of 3D imaging in pathology remains challenging. 3D volumetric datasets are orders of magnitude larger than standard 2D datasets, requiring substantially higher cost for acquisition, storage, and manual volumetric review and annotation. More fundamentally, the vast majority of clinical archives and widely used benchmarks remain 2D only, with billions of slides representing decades of medical history inaccessible to methods that explicitly exploit z-axis morphological information. While recent efforts have attempted to reconstruct volumes through the registration of serial biopsy samples \cite{redekop2025generating} or GAN-based single-cell synthesis \cite{gan_single_cell}, these methods either require specialized tissue preparation or lack the capacity to model the complex, multi-cellular architectural changes observed in whole-tissue pathology.

Herein, we propose PathoS2VD (Pathology Slice-to-Volume Diffusion), a generative framework for learning transferable z-axis morphological priors from pathology z-stack data. Instead of reconstructing true tissue volumes, PathoS2VD synthesizes plausible pseudo z-stacks that enrich conventional 2D pathology images with complementary cross-plane morphological information. To enable deployment on large-scale clinical datasets without native z-stack acquisition, we further introduce a prior-preserving mixture fine-tuning strategy that transfers the learned z-axis before conventional 2D pathology domains. Experiments on one z-stack dataset and two downstream prostate datasets demonstrate that the generated pseudo z-stacks exhibit realistic cross-plane morphology and consistently improve both patch-level and slide-level cancer classification over 2D baselines.

\section{Related Method}

Computational pathology is often hindered by data scarcity and poor domain generalization. Generative models address these limitations by synthesizing high-fidelity training instances. While early approaches relied on VAEs and GANs, Diffusion Models (DMs) have emerged as the state-of-the-art, generating images via iterative denoising \cite{pathogen}. Offering superior semantic fidelity and training stability, DMs are increasingly adopted for complex pathology tasks, including report-guided generation, gigapixel WSI synthesis, and rare-label augmentation.

\subsection{Video Diffusion for Z-axis Morphology Modeling} 
Compared with a single tissue section, a z-stack captures complementary morphological information across adjacent focal planes. While native 3D imaging techniques such as OTLS and MicroCT provide this information directly, their clinical adoption remains limited, and reconstruction from serial tissue sections requires specialized acquisition protocols that are unavailable in routine pathology \cite{redekop2025generating,valis}.

Rather than reconstructing physical tissue volumes, our objective is to learn the gradual morphological variation across ordered focal planes. This setting differs from conventional 3D scene reconstruction, where multiple viewpoints are available, such as NeRF \cite{nerf} or Gaussian Splatting \cite{gaussian_splatting}, but naturally resembles video generation, where neighboring frames exhibit strong local correlations. We therefore adopt the Latent Video Diffusion Model (LVDM) \cite{lvdm, svd} as the backbone and reinterpret the temporal dimension as the z-axis of a pathology z-stack. The spatial-temporal attention mechanism enables the model to capture cross-plane morphological dependencies, which are subsequently transferred to conventional 2D pathology through pseudo z-stack synthesis.

\section{Preliminaries: Latent Diffusion Model}
This section reviews latent diffusion models (LDMs) \cite{ldm}, which form the basis of LVDMs \cite{lvdm}. LDMs perform DMs in the latent space of a pretrained VAE.

Let $\mathbf{x} \in \mathbb{R}^{H\times W \times 3}$ be a RGB image. The VAE encoder $\mathcal{V}_{E}(\cdot)$ encodes $\mathbf{x}$ into a latent representation $\mathbf{z} = \mathcal{V}_{E}(\mathbf{x}) \in \mathbb{R} ^{h\times w\times 4}$, where $f = H/h = W/w$ is the downsampling factor. The VAE decoder $\mathcal{V}_{D}(\cdot)$ is trained to reconstruct the input $\mathbf{x}$ from the latent representation $\mathbf{z}$ such that $\hat{\mathbf{x}} = \mathcal{V}_{D}(\mathbf{z}) \approx \mathbf{x}$. 
 
Subsequently, DMs learn the latent data distribution $p_{data}(\mathbf{z})$ by defining the forward and reverse processes. While \cite{ddpm} designs the generative process as a discrete-time random process and score-based models \cite{score_dm} formulates it as stochastic differential equations (SDEs), EDM \cite{edm} provides a common framework that supports customized noisy schedules parameterized by a noise scale $\sigma_t$. In the forward process $\{\mathbf{z}_t\}_{t\in[0, T]}$, the clean data $\mathbf{z}_0$ is perturbed with Gaussian noise of varying noise scales $\sigma_t$:
 
\begin{equation}
\mathbf{z}_t = \mathbf{z}_0 + \sigma_t \epsilon, \quad \text{where } \epsilon \sim \mathcal{N}(\mathbf{0}, \mathbf{I}).
\end{equation}
The generation process is modeled as a reverse probability flow ordinary differential equation (PF-ODE), which establishes a deterministic mapping from the noise distributions to the data distribution. The EDM ODE is given by:
\begin{equation}
\label{eq: ode}
    d\mathbf{z}_t = \left[ -\sigma_t \nabla_{\mathbf{z}_t} \log p(\mathbf{z}_t; \sigma_t) \right] dt
\end{equation}

\noindent where $\nabla_{\mathbf{z}_t} \log p(\mathbf{z}_t; \sigma_t)$ is the score function, representing the gradient of the log-likelihood with respect to data. To solve this, a time-dependent denoising network $\mathcal{F}_{\theta}(\mathbf{z}_t; \sigma_t)$ is trained to estimate the clean data $\mathbf{z}_0$ from a noisy data $\mathbf{z}_t$. The training objective minimizes $L_2$ denoising error:
\begin{equation} 
\mathbb{E}_{\sigma, \mathbf{z}_0} \left[ \lambda(\sigma) \left\| \mathcal{F}_{\theta}(\mathbf{z}; \sigma) - \mathbf{z} \right\|_2^2 \right]
\end{equation}
where $\lambda(\sigma)$ is a weighting function. \cite{edm} shows that representing the denoiser $\mathcal{F}_\theta$ with a noise predictor $\epsilon_\theta (\cdot)$ yields better performance:
\begin{equation}
\mathcal{F}_{\theta}(\mathbf{z},\sigma) = c_{\text{skip}}(\sigma)\, \mathbf{z} + c_{\text{out}}(\sigma)\, \epsilon_{\theta}(c_{\text{in}}(\sigma)\, \mathbf{z}; c_{\text{noise}}(\sigma))
\end{equation}
where $c_{\text{skip}}$, $c_{\text{out}}$, $c_{\text{in}}$, and $c_{\text{noise}}$ are EDM scaling functions, for which we use standard EDM hyperparameters. Finally, the score function can be expressed by $\mathcal{F}_\theta$ as $\nabla_{\mathbf{z}_t} \log p(\mathbf{z}_t; \sigma_t) = \frac{\mathcal{F}_{\theta}(\mathbf{z}_t; \sigma_t) - \mathbf{z}_t}{\sigma_t^2}$. Once trained, samples are generated by initializing $\mathbf{z}_T$ from noise and numerically integrating Eq.(\ref{eq: ode}) on $t = T \rightarrow 0$.

\section{Method}
We leverage two types of datasets. The first is a prostate pathology z-stack dataset $\mathbf{\mathcal{D}_{zstack}}$ \textbf{(source)}, used to learn z-axis prior from a 2D observation. 
The second is a 2D cancer classification dataset $\mathbf{\mathcal{D}_{cancer}}$ \textbf{(target)}, which is used for domain adaptation and downstream clinical evaluation of the generated z-stacks.
In this section, we present the curation of the z-stack data, our two-stage training strategy, designed to address the spatial domain shift between the source and target datasets, and inference strategy for volume generation and expansion.

\begin{figure*}[t]
    \centering
    \includegraphics[width=1\textwidth]{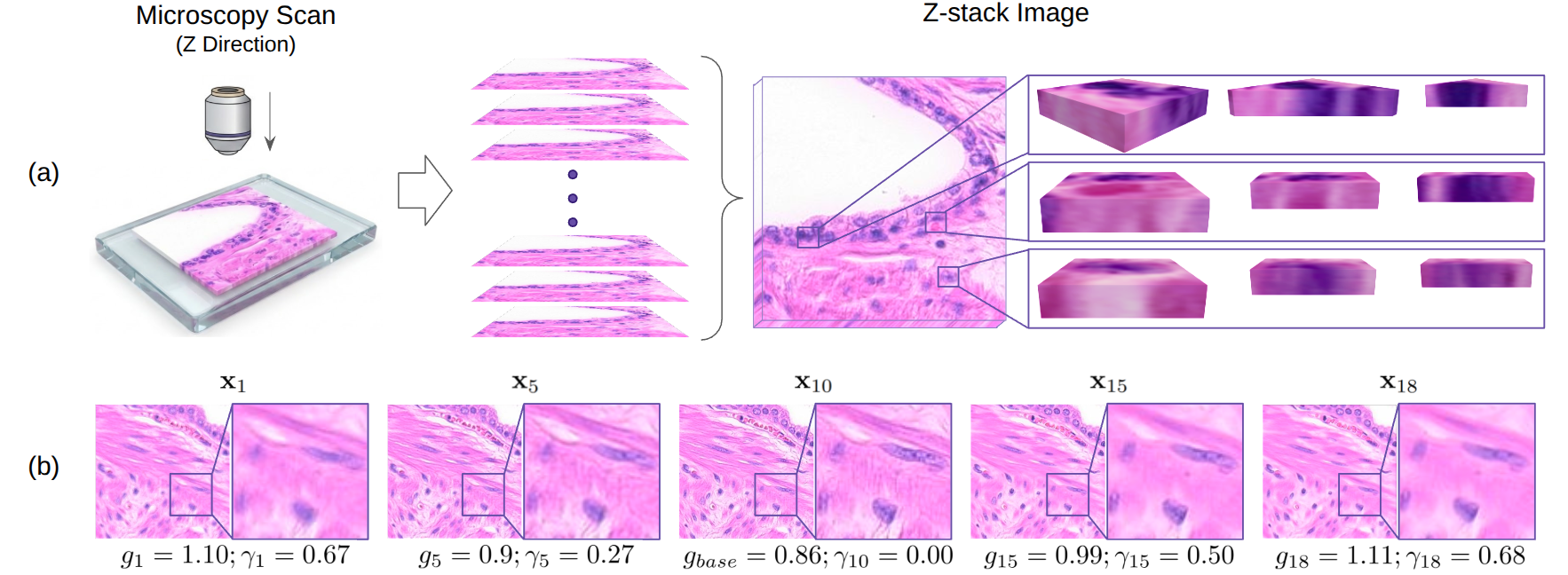}
    \caption{(a) A z-stack image is a sequence of slices that capture structural details across depths. (b) $\{x_k\}_{k=1,5,10,15,18}$ from a raw z-stack sample with their absolute blur $g_k$ and relative blur $\gamma_k$. $x_{10}$ is chosen as the focal plane as $\gamma_{10} = 0$ in this example.}
    \label{fig:fig0}
\end{figure*}

\subsection{Z-stack Dataset Curation}
\label{sec:data_curation}
Z-stack imaging captures a sequence of slices/frames at discrete focal depths, which is often used to overcome the limitations of shallow depth of field in microscopy. By combining multiple focal plane slices, z-stack imaging can retain structural details across different depths, as in Fig. \ref{fig:fig0} (a). In our study, we propose to leverage z-stack imaging to learn z-axis morphological priors and synthesize pseudo z-stack representations from a single 2D image.

Let $\{\mathbf{X}_{raw}^{(i)}\}_{i=1}^{N_s}$ denote a collection of $N_s$ raw z-stack tissue samples. Each volume $\mathbf{X}_{raw}^{(i)}=\{\mathbf{x}^{(i)}_{k}\}_{k=1}^{N_{o}}$ comprises $N_o$ RGB slices $\mathbf{x}^{(i)}_{raw,k} \in \mathbb{R}^{H \times W \times 3}$, corresponding to different focal planes of a tissue section, where $N_o=19$ and slices are acquired at a focal interval of 0.3 $\mu$m. While these z-stack volumes provide rich cross-plane morphological information, the individual slices $\mathbf{x}^{(i)}_{raw,k}$ are not uniformly sharp. Slices that are distal from the optimal focal plane exhibit a varying degree of defocus blur (Fig. \ref{fig:fig0} (b)), which is detrimental to the accurate representation of tissue morphology.
 
To account for defocus blur, we adopt a physical defocus model where a blurred image $\mathcal{I}_b$ is the convolution of a sharp image $\mathcal{I}$ with a Gaussian point spread function (PSF) $G(g_k)$, where $g_k$ represents the blur magnitude at depth $k$: $\mathcal{I}_b = \mathcal{I} \ast G(g_k)$.
Traditional blur estimation methods, such as Laplacian variance \cite{pech2000diatom}, are insufficient, as they are content-dependent, in which textured tissue regions yield higher blur scores than smooth regions, regardless of the blur degree.

To address this, we formulate a relative blur metric $\gamma^{(i)}_k$ under two key assumptions: 1) all slices $\mathbf{x}_{raw,k}^{(i)}$ within the $i$th volume share the same "true" content $\mathcal{I}^{(i)}$ and 2) the sharpest slice in the volume is formulated as $\mathcal{I}^{(i)} \ast G(g^{(i)}_{base})$, and other slices result from applying an additional relative blur kernel $\gamma^{(i)}_k$, defined as: 
% TODO
\begin{equation} 
\mathbf{x}^{(i)}_{raw,k} \approx \mathcal{I}^{(i)} \ast G(g^{(i)}_{base}) \ast G(\gamma^{(i)}_k)
\end{equation} 
where $g^{(i)}_{base} = \min_k (g^{(i)}_k)$ is the blur score of the sharpest slice within the volume (Fig. \ref{fig:fig0} (b)). Following \cite{defocus}, we calculate the absolute blur degree $g_{k}$ for each slice. Using the additive property of Gaussian variances under convolution, $(g_k)^2 = (g_{base})^2 + (\gamma_k)^2$, we derive the relative blur parameter:
\begin{equation}
\gamma^{(i)}_k = \sqrt{\big( g^{(i)}_k \big)^2 - \big( g^{(i)}_{base} \big)^2}.
\end{equation}
Accordingly, the sharpest slice with $\gamma^{(i)}_k=0$ corresponds to the optimal focal plane, and the higher the value of $\gamma^{(i)}_k$, the greater degree of blur it possesses relative to the sharpest slice in the $i$th volume.

%For the $\mathcal{D}_{zstack}$ construction,
We utilize the relative blur metric $\gamma^{(i)}_k=0$ to perform focal plane alignment. Specifically, we shift the slice indices in each volume $\mathbf{X}^{(i)}_{raw}$ such that the slice corresponding to the optimal focal plane, where $\gamma_{k} = 0$, is centered in the middle of the volume. After shifting, the slice at index $k=(N_o + 1)/2$ has $\gamma_{k} = 0$. If the focal plane is near the volume boundary, this shift creates empty slots, which we fill with pseudo-slices containing infinite blur ($\gamma_k \to \infty$). Trimming and padding all volumes to a fixed depth of $N_o = 11$, corresponding to a $\pm$1.5$\mu$m tissue context, we construct a 3D z-stack dataset, designated as:
\begin{equation}
\mathcal{D}_{zstack} = \{\mathbf{X}^{(i)}\}_{i=1}^{N_s}, \quad \mathbf{X}^{(i)}=\{\mathbf{x}^{(i)}_{k}\}_{k=1}^{N_o}
\end{equation}
where $\mathbf{X}^{(i)}$ denotes the refined volume and $\mathbf{x}^{(i)}_{k}$ represents refined individual slices within the volume.

\subsection{Pseudo Z-stack Generation}
\label{sec:inference}
PathoS2VD is built upon LVDM \cite{lvdm} that augments a standard 2D U-Net backbone, parameterized by $\theta$, with temporal modules that are parameterized by $\phi$. The temporal modules include 3D convolutional blocks and temporal attention layers, designed to capture dependencies across the video frames. In our settings, we reinterpret the temporal dimension as the ordered focal-plane dimension of a pathology z-stack.  Thus, the image backbone $\theta$ processes each slice as an independent image, while the temporal layers $\phi$ treat them as a sequence with depth structural correlations along the tissue depth.

Given a 2D image $\mathbf{y}^{(i)}\in \mathbb{R}^{H \times W \times 3}$, PathoS2VD aims to synthesize a pseudo z-stack tissue $\mathbf{X}^{(i)} = \{\mathbf{x}_{k}^{(i)}\}_{k=1}^{N_o}$ that preserves the tissue morphology while introducing plausible z-axis variations. Given a pretrained sampler $p_{\theta, \phi}(\mathbf{Z}_{t-1}|\mathbf{Z}_t)$ and semantic guidance $\mathbf{c}^{(i)} = CLIP(\mathbf{y}^{(i)})$, $\mathbf{X}^{(i)}$ is obtained by decoding a latent clean sample $\mathbf{Z}_0$ that iteratively denoised from a sampled Gaussian Noise $\mathbf{Z}_T$:
\begin{equation}
\mathbf{X}^{(i)} = \mathcal{V}_D (\mathbf{Z}_0),  \mathbf{Z}_0 \sim p_{\theta, \phi}(\mathbf{Z}_0 | \mathbf{Z}_{1\sim T}, \mathbf{c}^{(i)})
\end{equation}

% $\mathbf{Z}_0, \mathbf{Z}_t, \hat{\mathbf{Z}}_0, \mathbf{Z}_T$
% $\mathcal{V}_E, \mathcal{V}_D$
With $N_o=11$, our generative model extends an image into a tissue of $\pm$1.5$\mu$m thickness, defined as: $G_{3\mu m}: \mathbb{R}^{H\times W \times 3} \rightarrow \mathbb{R}^{N_o\times H \times W \times 3}$, i.e., $\mathbf{X}^{(i)} = G_{3\mu m}(\mathbf{y}^{(i)}) = [\mathbf{x}^{(i)}_1, \mathbf{x}^{(i)}_2, \dots, \mathbf{x}^{(i)}_{N_o}]$, where input image $\mathbf{y}^{(i)}$ corresponds to the focal plane of $\mathbf{X}^{(i)}$.

\subsection{Efficient Pseudo Z-stack Expansion}
\label{sec:expansion}
To generate a thicker tissue context beyond $\pm 1.5\mu m$, we propose an iterative outpainting strategy that extends PathoS2VD with a volume-expansion generator: $G_{6\mu m}: \mathbb{R}^{H\times W \times 3} \rightarrow \mathbb{R}^{2N_o\times H \times W \times 3}$. The key idea is to extrapolate along the depth axis in both directions by reusing the boundary slices of the $3\mu$m-thick pseudo volume as conditioning inputs for subsequent upward and downward generations, respectively.
Specifically, given $\mathbf{X}^{center}$, the top slice $\mathbf{x}_1^{center}$ and bottom slice $\mathbf{x}_{N_o}^{center}$ serve as the conditioning inputs $\mathbf{c}^{up}$ and $\mathbf{c}^{down}$, respectively.
 
Sequentially generating up and down volumes introduces substantial computational burden. Using standard classifier-free guidance (CFG) \cite{cfg}, it requires two U-Net evaluations at each timestep $t$ (conditional $\mathbf{c}$ and unconditional $\emptyset$): $\tilde{\epsilon}_t = w \epsilon_{\theta, \phi}(\mathbf{z}_t, \mathbf{c}) + (1-w) \epsilon_{\theta, \phi}(\mathbf{z}_t, \emptyset)$, where $w$ is CFG guidance weight.
Hence, generating three z-stacks ($\mathbf{X}^{up}, \mathbf{X}^{center}, \mathbf{X}^{down}$) requires 6 evaluations per timestep.

To mitigate this, we propose a shared Null-space acceleration strategy. Our key assumption is that adjacent tissue blocks share similar structural distributions (stain, cell density, background texture), implying that the unconditional score estimate $\epsilon_{\theta, \phi}(\mathbf{z}_t, \emptyset)$ remains stable across locally adjacent volume blocks.
Under this assumption, we fix the initial random noise seed $\mathbf{Z}_T$ across all three z-stack generations. During the inference of $\mathbf{X}^{center}$, we cache the unconditional noise prediction $\mathcal{M} = \{\epsilon^\emptyset_t = \epsilon_{\theta, \phi}(\mathbf{Z}_t^{center}, \emptyset) \}_{t = 1}^{T}$. For the peripheral volumes, we bypass the unconditional forward pass and reuse the cached $\epsilon^\emptyset_t$:
\begin{equation}
    \tilde{\epsilon}^{k}_t = w \epsilon_{\theta, \phi}(\mathbf{Z}_t^{k}, \mathbf{c}^k) + (1-w)\epsilon^\emptyset_t
\end{equation}
where $k \in \{up, down\}$. This reduces the computational cost from 6 to 4 evaluations per step, providing a 33$\%$ speedup. The complete procedure is detailed in Alg. \ref{alg:inference}.

% The generated upward and downward pseudo z-stacks overlap with the central pseudo z-stack. A straightforward strategy is to construct a seamless expanded z-stack by stitching the overlapping regions:
% \begin{equation}
%     \{\mathbf{x}_k^{up}\}_{k=\frac{D}{2}}^D
%     =
%     \{\mathbf{x}_k^{center}\}_{k=1}^{\frac{D}{2}}, \quad
%     \{\mathbf{x}_k^{down}\}_{k=1}^{\frac{D}{2}}
%     =
%     \{\mathbf{x}_k^{center}\}_{k=\frac{D}{2}}^D.
% \end{equation}
% This approach removes duplicated slices and produces a single continuous pseudo z-stack. However, due to the stochastic nature of diffusion sampling, the overlapping regions are not pixel-wise identical, even though they represent the same tissue context. Enforcing hard stitching therefore discards one of two plausible morphological realizations and may remove useful complementary information.

% Instead, we retain all generated slices from $\mathbf{X}^{up}$, $\mathbf{X}^{center}$, and $\mathbf{X}^{down}$, and regard the overlapping regions as multiple plausible observations of the same transition zone. This effectively serves as a form of test-time augmentation, improving the robustness of downstream feature aggregation against generation uncertainty. 

The generated upward and downward pseudo z-stacks overlap with the central pseudo z-stack. One option is to stitch the overlapping regions to construct a continuous expanded z-stack:
\begin{equation}
    \{\mathbf{x}_k^{up}\}_{k=\frac{D}{2}}^D
    =
    \{\mathbf{x}_k^{center}\}_{k=1}^{\frac{D}{2}}, \quad
    \{\mathbf{x}_k^{down}\}_{k=1}^{\frac{D}{2}}
    =
    \{\mathbf{x}_k^{center}\}_{k=\frac{D}{2}}^D.
\end{equation}
Alternatively, all generated slices can be retained, treating the overlapping regions as multiple plausible realizations of the same tissue context. The choice depends on the downstream application, and both strategies are supported by the proposed inference framework.

\begin{algorithm}[tb]
   \caption{Expansion with Shared Null-space}
   \label{alg:inference}
\begin{algorithmic}[1]
   \STATE {\bfseries Input:} Image $\mathbf{y}$, Diffusion Steps $T$, CFG Scale $w$, Model $(\epsilon_{\theta, \phi}, \mathcal{F}_{\theta, \phi})$.
   \STATE {\bfseries Output:} Bag of Instances $\mathcal{B}$.
   
   \STATE Extract embedding $\mathbf{c}^{center} = \text{CLIP}(\mathbf{y})$.
   \STATE Sample initial noise $\mathbf{Z}_T^{center} = \mathbf{Z}_T \sim \mathcal{N}(\mathbf{0}, \mathbf{I})$. 
   \STATE Initialize empty cache $\mathcal{M} = \{\}$.

% 2. Phase I: Center Volume Generation (with Caching)
   \FOR{$t = T$ {\bfseries to} $1$}
       \STATE $\epsilon^\emptyset_t \leftarrow \epsilon_{\theta, \phi}(\mathbf{Z}_t^{center}, \emptyset, t)$ \COMMENT{Unconditional pass}
       \STATE $\epsilon^{cond}_t \leftarrow \epsilon_{\theta, \phi}(\mathbf{Z}_t^{center}, \mathbf{c}^{center}, t)$ \COMMENT{Conditional pass}
       \STATE $\tilde{\epsilon}_t \leftarrow \epsilon^\emptyset_t + w \cdot (\epsilon^{cond}_t - \epsilon^\emptyset_t)$ \COMMENT{CFG}
       \STATE $\mathbf{Z}_{t-1}^{center} \leftarrow \text{ODE sampler}(\mathbf{Z}_t^{center}, \tilde{\epsilon}_t)$
       \STATE $\mathcal{M}[t] \leftarrow \epsilon^\emptyset_t$ \COMMENT{Cache null-conditional noise}
   \ENDFOR
   \STATE Decode $\mathbf{X}^{center}=[\mathbf{x}_1,\dots, \mathbf{x}_D] \leftarrow \mathcal{D}(\mathbf{Z}_0^{center})$.
   
    % Prepare Boundary Conditions}
   \STATE $\mathbf{c}^{up} \leftarrow \text{CLIP}(\mathbf{x}_1), \mathbf{c}^{down} \leftarrow \text{CLIP}(\mathbf{x}_D)$.
   \STATE $\mathbf{Z}_T^{up} = \mathbf{Z}_T^{down} = \mathbf{Z}_T$. \COMMENT{Reuse initial noise}

   % Boundary Extension (Accelerated)
   \FOR{$t = T$ {\bfseries to} $1$}
       \STATE Retrieve $\epsilon^\emptyset_t \leftarrow \mathcal{M}[t]$. \COMMENT{Reuse cache}
       \FOR{$k \in \{up, down\}$}
           \STATE $\epsilon^{cond}_t \leftarrow \epsilon_{\theta, \phi}(\mathbf{Z}_t^{k}, \mathbf{c}^{k}, t)$
           \STATE $\tilde{\epsilon}_t^{k} \leftarrow \epsilon^\emptyset_t + w \cdot (\epsilon^{cond}_t - \epsilon^\emptyset_t)$
           \STATE $\mathbf{Z}_{t-1}^{k} \leftarrow \text{ODE sampler}(\mathbf{Z}_t^{k}, \tilde{\epsilon}_t^{k})$
       \ENDFOR
   \ENDFOR
   
   \STATE Decode $\mathbf{X}^{up} \leftarrow \mathcal{D}(\mathbf{Z}_0^{up})$, $\mathbf{X}^{down} \leftarrow \mathcal{D}(\mathbf{Z}_0^{down})$.
   \STATE Construct Bag $\mathcal{B} = \mathbf{X}^{up} \cup \mathbf{X}^{center} \cup \mathbf{X}^{down}$.
   \STATE {\bfseries return} $\mathcal{B}$
\end{algorithmic}
\end{algorithm}

\subsection{Blur-aware Masking Z-stack Training}
\label{sec:stage_I}

We train PathoS2VD on $\mathcal{D}_{zstack}$ to learn cross-plane morphological correlations. The VAE is fine-tuned following the standard objective in \cite{ldm}. Since slices farther from the focal plane suffer from severe defocus effect, directly supervising all the slices may encourage the model to reproduce the defocus effect instead of tissue morphology. \ To mitigate this issue, we introduce a blur-aware masking strategy, which utilizes the relative blur metric $\gamma_k^{(i)}$ (Section \ref{sec:data_curation}). For each z-stack $\mathbf{X}^{(i)}$, we define a binary validity mask $M^{(i)}\in \{0, 1\}^{N_o}$ by setting a global blur threshold $\tau_{blur}$ (empirically set to $0.2$). The validity mask for slice $k$ in volume $\mathbf{X}^{(i)}$ is defined as:
\begin{equation}
M^{(i)}_k = \mathds{1}(\gamma^{(i)}_k < \tau_{blur}).
\end{equation}
where $\mathds{1}$ is the indicator function. Any padding/pseudo slices are naturally excluded by setting($\gamma \to \infty$). 

To adopt plane-guided volume reconstruction, we use the image embedding $\mathbf{c}^{(i)}$ of the focal plane in $\mathbf{X}^{(i)}$ as the condition. The training objective for volume reconstruction in the latent space  $\mathbf{Z}^{(i)} = \mathcal{V}_E(\mathbf{X}^{(i)})$ can be written as:
\begin{equation}
\label{blur_loss}
    \mathcal{L}_{blur-aware} = \mathbb{E}_{i, \sigma} [\lambda(\sigma)  M^{(i)} \left\|\mathcal{F}_{\theta, \phi}(\mathbf{Z}^{(i)}; \sigma, \mathbf{c}^{(i)}) - \mathbf{Z}^{(i)} \right\|_2^2]
\end{equation}
This ensures the generated z-stack is symmetrically aligned with the input image, and the model focuses on learning 3D tissue morphology from the z-stack data, effectively ignoring optical aberrations.

\subsection{Mixture Fine-tuning for Domain Adaptation (MFDA)}
\label{sec:stage_II}
We denote the z-stack dataset as the source dataset $D_{zstack}$ and the target image cancer dataset as $\mathcal{D}_{cancer} = \{ \mathbf{y} ^{(j)}\}_{j=1}^{N_t}$, which contains $N_t$ number of RGB images $\mathbf{y} ^{(j)} \in \mathbb{R}^{H \times W \times 3}$. Although $D_{stack}$ provides PathoS2VD with pathology structural priors, its appearance distribution differs from $D_{cancer}$. Directly applying the pretrained model often results in domain-shift artifacts (e.g., stain mismatch). 

Using the static volume constructed by duplicating a target image $N_o$ times collapses the depth-wise structural variations. Another option is to fine-tune the spatial layers in $\theta$ on target images as in \cite{videocrafter2}; however, we observed that this led to temporal instability, i.e., slice-to-slice inconsistency (flicker), likely because image-only supervision disrupts temporal dynamics learned during z-stack training (Section \ref{sec:stage_I}).

Following the joint image-video optimization paradigm adopted in recent video generation models, we adopt a Mixture Fine-tuning strategy for Domain Adaptation (MFDA) that smoothly transfers the model to the target image distribution. Unlike previous methods, our objective is not to improve video generation quality but to preserve the learned z-axis morphological prior while adapting the spatial appearance to conventional pathology datasets. 

We construct training batches comprising two streams: 1) Z-stack streams ($\mathbf{X} \in \mathcal{D}_{zstack}$); 2) Target image streams ($\mathbf{y} \in \mathcal{D}_{cancer}$).
The z-stack stream is optimized using the blur-aware objective in Eq.\ref{blur_loss}, preserving cross-plane morphological consistency. For the target image stream, only the spatial backbone $\theta$ is updated by bypassing the temporal modules $\phi$, allowing the model to adapt to the target appearance without disrupting the learned z-axis correlations. The key insight is to uncouple the updates but optimize the parameters jointly for 2D and 3D modules. Similar to Stage I, the VAE is first fine-tuned on the mixed data before optimizing the diffusion model. Then, for the LDM, we use image embedding $\mathbf{c}^{(j)}$ of $\mathbf{y}^{(j)}$ as an image condition to guide the generation. The training objective is a standard image diffusion loss applied to independent images in the latent space $\mathbf{z}^{(j)} = \mathcal{E}(\mathbf{y}^{(j)})$: 
\begin{equation}
    \mathcal{L}_{spatial-adapt} = \mathbb{E}_{j, \sigma} [\lambda(\sigma) \left\|\mathcal{F}_{\theta}(\mathbf{z}^{(j)}; \sigma, \mathbf{c}^{(j)}) - \mathbf{z}^{(j)} \right\|_2^2
\end{equation}
The joint objective is $\mathcal{L} = \mathcal{L}_{blur-aware} + \mathcal{L}_{spatial-adapt}$. MFDA enables the spatial backbone to adapt to the appearance distribution of conventional pathology datasets while maintaining the z-axis morphological prior learned from the source z-stack data.

\section{Experiment}

% \subsection{Datasets} 
% $\mathcal{D}_{zstack}$ includes a total of 80 WSIs, scanned with 19 focal planes spaced at a $0.3 \mu$m interval (representing a total physical depth of approximately $5.7 \mu$m) using the VISQUE DPS LH510 system (Vieworks Co., Ltd.), a high-throughput bright field scanner equipped with a 3-camera Realtime Extended Focus structure. From these WSIs, we extract 1,584,252 volume patches and compute the base blur $g_{base}$ for each patch as described in section \ref{sec:data_curation}. To ensure high data quality, we exclude patches where $g_{base}$ deviates by more than two standard deviations from the population mean, yielding $\mathcal{D}_{zstack}$ with 592,496 volume patches, serving as the source for learning 3D morphological priors.

% For the target dataset $\mathcal{D}_{cancer}$, we utilize the AGGC2022 Prostate Cancer Grading dataset \cite{aggc}, annotated with five categories: [Background, Benign, Gleason Grade 3 (G3), Gleason Grade 4 (G4), Gleason Grade 5 (G5)]. This dataset is composed of 3 subsets: images in subsets 1 and 2 are scanned by an Akoya Biosciences scanner, and subset 3 images are scanned by 6 scanners (Akoya Biosciences, Olympus, Zeiss, Leica, KFBio, and Philips). We use subsets 1 and 2 as training (87,720 patches) and validation (48,378 patches) datasets, and subset 3 as a testing dataset (187,595 patches). 

\subsection{Datasets}

Our experiments involve one source $\mathcal{D}_{zstack}$ and two downstream 2D prostate datasets $\mathcal{D}_{target}$. $\mathcal{D}_{zstack}$ consists of 80 whole-slide images (WSIs), each scanned at 19 focal planes with a spacing of $0.3~\mu$m (approximately $5.7~\mu$m total depth) using the VISQUE DPS LH510 system (Vieworks Co., Ltd.), a high-throughput bright-field scanner equipped with a three-camera Realtime Extended Focus module. From these WSIs, we extract 984,252 z-stack patches. Following Section~\ref{sec:data_curation}, the base blur $g_{base}$ is estimated for every patch, and samples whose $g_{base}$ deviates by more than two standard deviations from the population mean are discarded. This results in 592,496 high-quality z-stack patches, which serve as the source data for learning transferable z-axis morphological priors.

We evaluate the learned z-axis prior on two 2D prostate datasets. (i) \textbf{AGGC2022} \cite{aggc} is a prostate cancer grading dataset with five categories (Background, Benign (BN), Gleason Grade 3 (G3), Gleason Grade 4 (G4), and Gleason Grade 5 (G5)). Following the split in \cite{lee2025benchmarking}, subsets 1 and 2 are used for training (87,720 patches) and validation (48,378 patches), while subset 3, scanned by six different scanner vendors, is used for testing (187,595 patches), providing a challenging cross-scanner evaluation.
(ii) \textbf{SICAPv2} \cite{silva2020going} is a prostate whole-slide dataset containing 124 training WSIs and 31 testing WSIs. Following the official annotations, slide-level classification is performed over four tissue categories: BN, G3, G4, and G5.
Unless otherwise specified, all image patches are center-cropped to $256 \times 256$ pixels and scanned at $20\times$ magnification.

\subsection{Downstream Cancer Diagnostic Evaluation}
\label{sec:downstream_classification}

% To evaluate whether the learned z-axis prior provides complementary information for downstream diagnosis, we perform both patch-level and slide-level classification using frozen pathology foundation model features. Given a single input patch $\mathbf{y}^{(i)}\in D_{cancer}$, we obtain a semantic embedding $\mathbf{f}^{(i)}_0 = \mathcal{E}_{FM}(\mathbf{y}^{(i)}) \in \mathbb{R}^{D_{FM}}$. The synthesized pseudo z-stack $\mathbf{X}^{(i)} = \{\mathbf{x}_{k}^{(i)}\}_{k=1}^{N_o}$ are independently encoded into a bag of features $\mathcal{B}_f^{(i)} = \{\mathbf{f}_{k}^{(i)}\}_{k=1}^{N_o}$. The slice features are then aggregated by a feature fusion module to obtain a z-stack representation: $f_{stack}^{(i)} = fusion(\mathcal{B}_f^{(i)} )$, which is combined with the raw image feature by residual addition $f_{final}^{(i)} = f_0^{(i)} + f_{stack}^{(i)}$

% For patch-level classification, the final patch representation is fed directly into a linear classifier. The baseline uses only $\mathbf f_0^{(i)}$.
% For slide-level classification, each WSI is represented as a bag of patch features. Specifically, the fused patch representations from all patches within a slide are aggregated using ABMIL to obtain a slide-level representation, followed by a linear classifier. The 2D baseline follows the same procedure but replaces $\mathbf f_{final}^{(i)}$ with the raw feature $\mathbf f_0^{(i)}$.

To evaluate whether the learned z-axis prior provides complementary information for downstream diagnosis, we perform both patch-level and slide-level classification using frozen pathology foundation model features, and we use UNI \cite{uni} in our experiment. Given an input patch $\mathbf{y}^{(i)}\in\mathcal{D}_{target}$, we extract the raw feature
$\mathbf{f}_0^{(i)}=\mathcal{E}_{FM}(\mathbf{y}^{(i)})\in\mathbb{R}^{D_{FM}}$.
The synthesized pseudo z-stack
$\mathbf{X}^{(i)}=\{\mathbf{x}_k^{(i)}\}_{k=1}^{N_o}$
is encoded slice-by-slice, producing a set of slice features
$\mathcal{B}_f^{(i)}=\{\mathbf{f}_k^{(i)}\}_{k=1}^{N_o}$.
The slice features are then aggregated into a z-stack representation $\mathbf{f}_{stack}^{(i)} = Fusion(\mathcal{B}_f^{(i)})$, where $Fusion(\cdot)$ denotes a feature aggregation module. We evaluate several commonly used ensemble strategies, including Max Pool \cite{krizhevsky2012imagenet}, DSMIL \cite{li2021dual}, Self-Attention \cite{vaswani2017attention, clam}. The aggregated feature is finally combined with the raw image feature through residual addition, $\mathbf{f}_{final}^{(i)}=\mathbf{f}_0^{(i)}+\mathbf{f}_{stack}^{(i)}$.

For patch-level classification, the final patch representation $\mathbf{f}_{final}^{(i)}$ is fed into a 2-layer MLP classifier, while the 2D baseline uses only the raw feature $\mathbf{f}_0^{(i)}$. For slide-level classification, each whole-slide image is represented as a bag of fused patch features, which are further aggregated by ABMIL to obtain a slide-level representation before linear classification. The 2D baseline follows the same pipeline using only the raw patch features.

% \section{Results}
% \begin{table}[h]
%     \centering
%     \caption{\textbf{Quantitative Assessment of Z-stack Generation Fidelity.} We evaluate the reconstruction quality of the Z-stack dataset.}
%     \label{tab:zstack}
%     \resizebox{\columnwidth}{!}{
%     \begin{tabular}{|c|c|c|c|c|c|c|}
%         \hline
%         ~ & SSIM$\uparrow$ & LPIPS$\downarrow$ & SSIM$_v$$\downarrow$ & LPIPS$_v$$\downarrow$ & FID$\downarrow$ & FVD$\downarrow$ \\ 
%         \hline
%         Zero-shot & 0.3166 & 0.4876 & 0.4289 & 0.2807 & 131.24 & 1846.22 \\ 
%         \hline
%         Stage I & \textbf{0.6794} & \textbf{0.1797} & \textbf{0.0250} & \textbf{0.0273} & \textbf{13.34} & \textbf{31.71} \\ 
%         \hline
%     \end{tabular}
% }
% \end{table}

\section{Results}
% \begin{table}[h]
%     \centering
%     \caption{\textbf{Quantitative Assessment of Z-stack Generation Fidelity.} We evaluate the reconstruction quality of the generated volumes against the ground truth Z-stack dataset.}
%     \label{tab:zstack}
%     \resizebox{\columnwidth}{!}{
%     \begin{tabular}{ccccccc}
%         \hline
%         ~ & SSIM$\uparrow$ & LPIPS$\downarrow$ & FID$\downarrow$ & SSIM$_v$$\downarrow$ & LPIPS$_v$$\downarrow$ & FVD$\downarrow$ \\ 
%         \hline
%         SVD (Zero-shot) & 0.3166 & 0.4876 & 131.24 & 0.4289 & 0.2807 & 1846.22 \\ 
%         \hline
%         SVD & \textbf{0.6794} & \textbf{0.1797} & \textbf{13.34} & 0.0250 & 0.0273 & \textbf{31.71} \\ 
%         \hline
%         Pix2Pix & 0.6331 & 0.2436 & 14.20 & 0.0578 & 0.0715 & 140.59\\ 
%         \hline
%         OpenSoraPlan & 0.5410 & 0.3065 & 42.96 & \textbf{0.0119} & \textbf{0.0264} & 87.99 \\ 
%         \hline
%     \end{tabular}
% }
% \end{table}

\begin{table}[htbp]
\centering
\caption{\textbf{Quantitative Assessment of Z-stack Generation Fidelity.} We evaluate the reconstruction quality of the generated volumes against the ground truth Z-stack dataset.}
\label{tab:zstack} 

% This forces the table to scale perfectly down to the text width
\resizebox{\textwidth}{!}{
\begin{tabular}{l ccccccccc}
\toprule
\textbf{Method} & 
\makecell{SSIM\\$\uparrow$} & 
\makecell{LPIPS\\$\downarrow$} & 
\makecell{FID\\$\downarrow$} & 
\makecell{FVD\\$\downarrow$} & 
\makecell{Orientation\\Similarity $\uparrow$} & 
\makecell{ST\\Frobenius $\downarrow$} & 
\makecell{Anisotropy\\MAE $\downarrow$} & 
\makecell{Vz\\Wasserstein $\downarrow$} & 
\makecell{ST\\Evolution\textsubscript{corr} $\uparrow$} \\
\midrule
PathoS2VD          & \textbf{0.6794} & \textbf{0.1797} & \textbf{13.34} & \textbf{31.71} & \textbf{0.9407} & \textbf{0.7727} & \textbf{0.0306} & \textbf{0.0234} & \textbf{0.8949} \\
Pix2Pix      & 0.6331          & 0.2436          & 14.20          & 140.59         & 0.8612          & 257.5114        & 0.0567          & 0.0914          & 0.7707          \\
OpenSoraPlan & 0.5410          & 0.3065          & 42.96          & 87.99          & 0.9025          & 3.5136          & 0.0433          & 0.0505          & 0.8803         \\
\bottomrule
\end{tabular}
} % End of \resizebox
\end{table}

\begin{table}[t]
    \centering
    \caption{\textbf{Quantitative Assessment of Generation Fidelity.} We evaluate the reconstruction quality of the central slice in the generated volume against the input AGGC2022 image across different cancer grades. }
    \label{tab:aggc}
    \resizebox{\columnwidth}{!}{
    \setlength{\tabcolsep}{5pt}
    \begin{tabular}{lcccccccc}
        \toprule
        & & \multicolumn{6}{c}{Gleason Grade Class} \\
        \cmidrule(lr){3-8}
        % \textbf{Metric} & \textbf{Stage} & \textbf{Background} & \textbf{Benign} & \textbf{G3} & \textbf{G4} & \textbf{G5} & \textbf{Total} \\
        Metric& Stage & Background & Benign & G3 & G4 & G5 & Total\\
        \midrule
        \multirow{3}{*}{SSIM $\uparrow$} 
        & Zero-shot SVD  & 0.6682 & 0.7075 & 0.6644 & 0.6651 & 0.6693 & 0.6339 \\
        & PathoS2VD  & 0.8115 & 0.7884 & 0.7755 & 0.7803 & 0.7556 & 0.7823 \\
        & PathoS2VD w/ MFDA & \textbf{0.8249} & \textbf{0.7940} & \textbf{0.7879} & \textbf{0.7902} & \textbf{0.7607} & \textbf{0.7915} \\
        \midrule
        \multirow{3}{*}{LPIPS $\downarrow$} 
        & Zero-shot SVD  & 0.2498 & 0.2228 & 0.2598 & 0.2488 & 0.2466 & 0.2712 \\
        & PathoS2VD  & 0.1358 & 0.1475 & 0.1548 & 0.1530 & 0.1606 & 0.1504 \\
        & PathoS2VD w/ MFDA & \textbf{0.1250} & \textbf{0.1429} & \textbf{0.1397} & \textbf{0.1402} & \textbf{0.1541} & \textbf{0.1404} \\
        \bottomrule
    \end{tabular}
    }
\end{table}

% \begin{figure*}[t]
%     \centering
%     \includegraphics[width=1\textwidth]{fig/Fig3.png}
%     \caption{Qualitative visualization of generated 3 $\mu$m volumes. We display three representative slices ($x_1, x_5, x_{10}$) from two generated samples. Yellow boxes highlight specific morphological evolutions along the z-axis, including focus shifts (blur to sharp), cellular rotation, and shape stretching, demonstrating the model's ability to simulate realistic optical sectioning.}
%     \label{fig:fig3}
% \end{figure*}

% \begin{figure*}[t]
%     \centering
%     \includegraphics[width=1\textwidth]{fig/Fig4.png}
%     \caption{Extended 6 $\mu$m volume generation from Benign tissue (top) and Gleason Grade 5 (bottom). We display selected slices from the iteratively generated Top, Center, and Bottom blocks. The sequence demonstrates consistent anatomical continuity over a larger depth, capturing complex volumetric events such as cellular overlapping and significant focal plane transitions (scanner approach effect).}
%     \label{fig:fig4}
% \end{figure*}

\begin{figure*}[t]
    \centering
    \includegraphics[width=1\textwidth]{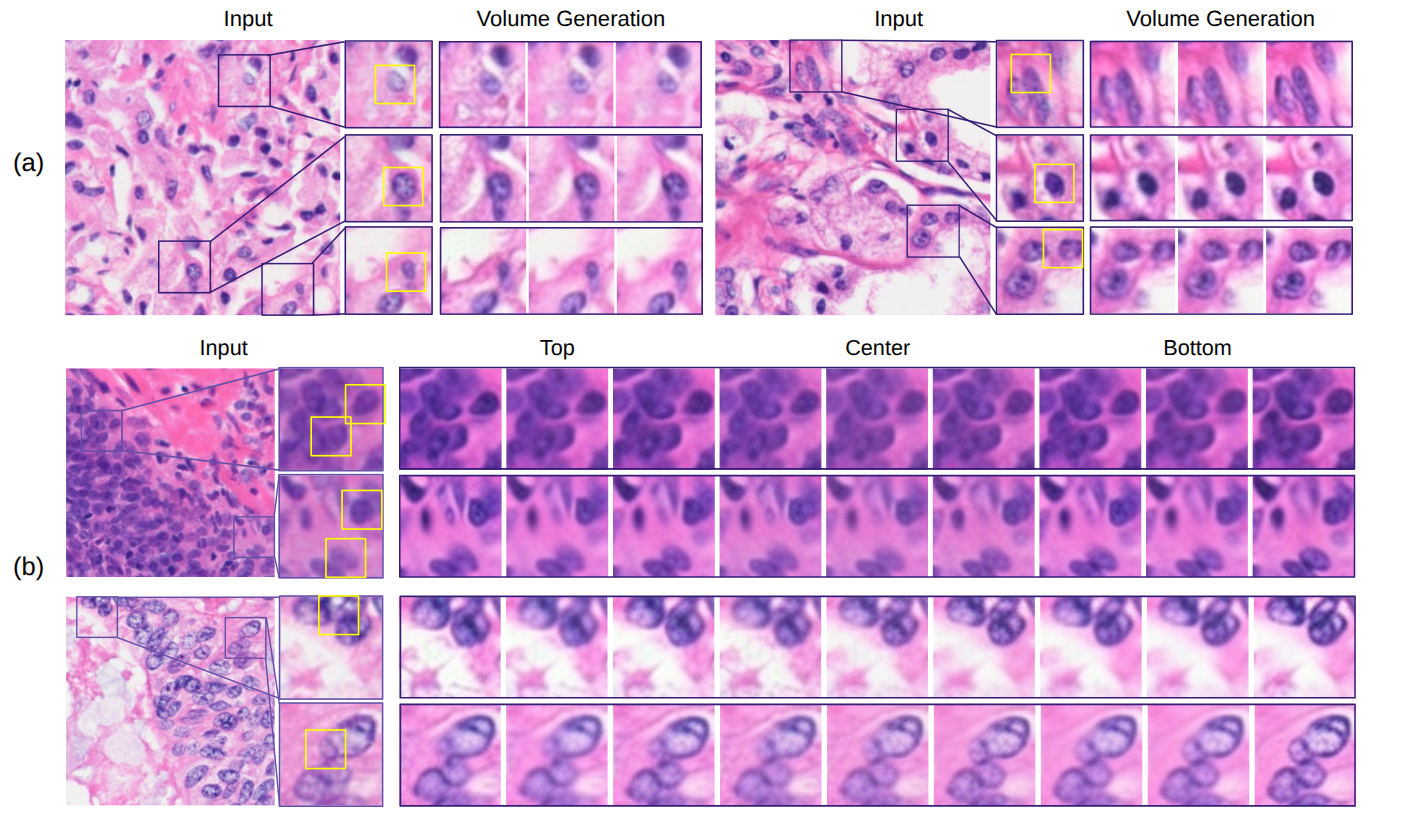}
    \caption{Qualitative visualization of generated volumes with three representative slices ($x_1, x_6, x_{11}$) for each volume. (a) Two samples of 3$\mu$m generated volumes, with yellow boxes highlighting specific morphological evolutions along the z-axis, including focus shifts (blur to sharp), cellular rotation, and shape stretching, demonstrating the model's ability to simulate realistic optical sectioning. (b) Top, Center, and Bottom blocks of extended 6$\mu$m volume generation from Benign tissue (top) and Gleason Grade 5 (bottom). The sequence demonstrates consistent anatomical continuity over a larger depth, capturing complex volumetric events such as cellular overlapping and significant focal plane transitions (scanner approach effect).}
    \label{fig:fig34}
\end{figure*}

\subsection{Quantitative Fidelity and Qualitative Analysis of Generated Pseudo Z-stacks}

We first evaluate the generation fidelity of PathoS2VD on the source z-stack dataset $\mathcal{D}_{zstack}$. Besides our SVD-based model, we compare against two representative baselines: the transformer-based video diffusion model OpenSoraPlan \cite{lin2024open} and the image-to-image GAN Pix2Pix \cite{isola2017image}. We report SSIM \cite{ssim}, LPIPS \cite{lpips}, FID \cite{fid}, and FVD \cite{fvd} to evaluate image and volume fidelity. To further assess biologically meaningful cross-plane morphology, we additionally adopt 3D structure tensor (ST)-based metrics, including Orientation Similarity, ST Frobenius, Anisotropy MAE, Vz Wasserstein, and ST Evolution$_{corr}$.

To quantify the preservation of cross-plane morphology, we further introduce structure tensor (ST)-based metrics. Specifically, we compute the 3D structure tensor for each z-stack and compare both local orientation and anisotropic tissue organization. Orientation Similarity measures the cosine similarity between the principal eigenvectors of paired structure tensors, while ST Frobenius measures the distance between the corresponding structure tensors. Anisotropy MAE evaluates the consistency of local structural anisotropy. Finally, Vz Wasserstein and ST Evolution$_{corr}$ characterize the statistical evolution of tissue morphology along the z-axis by comparing the distributions of principal orientation components and inter-slice structural changes, respectively.

% As shown in Table~\ref{tab:zstack}, PathoS2VD consistently outperforms Pix2Pix and OpenSoraPlan across all image fidelity metrics. Compared with Pix2Pix, PathoS2VD improves SSIM from 0.6331 to 0.6794 (+7.3\%), while reducing LPIPS by 26.2\%, FID by 6.1\%, and FVD by 77.4\%. Compared with OpenSoraPlan, the improvements are substantially larger, with SSIM increasing by 25.6\% and FVD decreasing by 63.9\%. These results indicate that adapting a pretrained image-to-video diffusion model to pathology z-stack data is considerably more effective than directly applying a generic video generation architecture.

As shown in Table~\ref{tab:zstack}, PathoS2VD consistently outperforms the other two models across all evaluation metrics. Compared with Pix2Pix, PathoS2VD improves SSIM by 7.3\% while reducing LPIPS, FID, and FVD by 26.2\%, 6.1\%, and 77.4\%, respectively. Compared with OpenSoraPlan, the improvements are even larger, particularly in FVD (-63.9\%), indicating substantially better cross-plane consistency.

The ST-based metrics further demonstrate that PathoS2VD preserves biologically meaningful cross-plane morphology. Orientation Similarity reaches 0.9407, indicating that the dominant local structural orientations are well preserved throughout the synthesized z-stack. Meanwhile, the substantially lower ST Frobenius distance (0.77 versus 4.82 and 257.51) suggests that the local spatial organization of tissue closely matches the ground truth. Consistently, the lowest Anisotropy MAE indicates that the generated pseudo z-stacks preserve the directional coherence of tissue microstructures more faithfully than competing methods.
Unlike the previous metrics, which compare local structural similarity, Vz Wasserstein and ST Evolution$_{corr}$ evaluate the statistical evolution of morphology along the z-axis. PathoS2VD achieves the lowest Vz Wasserstein distance (0.0234) and the highest ST Evolution$_{corr}$ (0.8949), demonstrating that it more faithfully reproduces the gradual morphological transitions across neighboring focal planes. These observations suggest that the learned z-axis prior captures not only the appearance of individual slices but also their cross-plane morphological progression.

% It should be noted that OpenSoraPlan is originally designed for first-frame-conditioned video generation with a temporal compression ratio of 4. To adapt it to z-stack synthesis, we only replaced the first-frame conditioning with a middle-slice condition while leaving the remaining architecture unchanged. Such a minimal adaptation does not account for the symmetric nature of z-stack generation and introduces a much higher temporal compression than the one adopted in PathoS2VD (4 vs. 1), limiting its ability to model subtle cross-plane morphological variations. Overall, these results highlight the importance of z-stack generation-specific design. Although pathology z-stacks can be represented as ordered image sequences, successfully modeling tissue depth requires learning domain-specific cross-plane morphological priors rather than generic temporal dynamics. This explains why adapting an image-to-video diffusion model with pathology-specific supervision substantially improves both visual fidelity and biologically meaningful structural consistency.

It is worth noting that OpenSoraPlan is originally designed for first-frame-conditioned video generation with a temporal compression ratio of 4. In our experiments, we only replaced the first-frame condition with a middle-slice condition while leaving the remaining architecture unchanged. Consequently, its causal design and aggressive temporal compression are not well matched to the symmetric and fine-grained nature of pathology z-stacks. These observations highlight that z-stack synthesis requires task-specific modeling of cross-plane morphology rather than directly transferring generic video generation models.

% To assess the effectiveness of the proposed MFDA strategy, we quantified reconstruction SSIM and LPIPS between the input 2D target images and the central slice of the generated 3 $\mu$m volumes. 
% As presented in Table \ref{tab:aggc}, PathoS2VD, trained on $D_{stack}$, was substantially superior to the zero-shot SVD, which yielded poor fidelity (SSIM $\approx$ 0.63 and LPIPS $\approx$ 0.25). With the integration of MFDA, PathoS2VD further improved the generation quality by 1.18\% in SSIM and 6.65\% in LPIPS, suggesting that the proposed MFDA bridges the domain gap between the generic source tissue and the target prostate cancer dataset, harmonizing the stain and texture distributions. 
% We further observed a class-dependent performance gradient: benign and low-grade tissue classes (background, Benign, and G3) achieved higher fidelity scores compared to high-grade cancer classes (G4 and G5). We attribute this to the distribution of the source z-stack dataset, which is predominantly composed of benign or healthy tissue structures. Consequently, the model possesses stronger priors for benign morphologies.

To evaluate the proposed prior-preserving mixture fine-tuning strategy, we compare the generated central slice with the input target image. As shown in Table~\ref{tab:aggc}, MFDA further improves SSIM by 1.18\% and reduces LPIPS by 6.65\% over the zero-shot model, demonstrating the importance of bridging the appearance gap between the source z-stack data and the target pathology domain. We further observed a class-dependent performance gradient: benign and low-grade tissue classes (background, Benign, and G3) achieved higher fidelity scores compared to high-grade cancer classes (G4 and G5). We attribute this to the distribution of the source z-stack dataset, which is predominantly composed of benign or healthy tissue structures. Consequently, the model possesses stronger priors for benign morphologies.

To verify that PathoS2VD learns true volumetric morphology rather than simple 2D repetitions, we qualitatively inspect the generated pseudo z-stacks in Fig.~\ref{fig:fig34}(a) shows representative 3$\mu$m generations by visualizing slices $[x_1,x_6,x_{11}]$. The synthesized sequences exhibit the "slow motion" characteristic of microscopic optical sectioning. As highlighted by the yellow boxes, we observed morphological evolution along the depth: 1) Optical sectioning effects: the cell appears smaller and blurrier when distant from the focal plane (left row 1) or progressively becomes larger and sharper as the focal plane intersects the cell body (left row 3; right row 1 and 2) and 2) Non-linear deformations: cellular rotation (left row 2) and shape stretching (right row 3) indicate that the model has learned the complex 3D geometry of tissue structures.
Moreover, Fig. \ref{fig:fig34} (b) demonstrates the capability of our iterative expansion strategy across distinct cancer grades (benign vs G5). We visualized the top, center, and bottom blocks from each to span the full 6$\mu$m depth. The extended range revealed more pronounced cellular dynamics. We observed distinct through-focus optical effects where cell contours transition from complete defocus to sharp (row 2 bottom; row 4). Furthermore, the extended depth captured complex spatial relationships, such as cell occlusion (row 1 bottom; row 3) and cross-section variations (row 1 top; row 2 top). 

\begin{table*}[t]
\centering
\caption{Quantitative comparison of feature aggregation methods on patch-level (AGGC2022) and slide-level (SICAPv2) classification tasks. The best result for each dataset and metric is shown in bold.}
\label{tab:classification}
\resizebox{1.0\textwidth}{!}{

\begin{tabular}{l|ccc|ccc}
\toprule
\multirow{2}{*}{Method}
& \multicolumn{3}{c|}{\textbf{AGGC2022 (Patch-level)}}
& \multicolumn{3}{c}{\textbf{SICAPv2 (Slide-level)}}\\
\cmidrule(lr){2-4}
\cmidrule(lr){5-7}
& Acc & W-F1 & M-F1
& Acc & W-F1 & M-F1\\
\midrule

2D MLP Baseline
&0.8131$\pm$0.0008
&0.8146$\pm$0.0024
&0.6780$\pm$0.0015
&0.7188$\pm$0.0255
&0.6490$\pm$0.0267
&0.4501$\pm$0.0495
\\

Max Pool
&\textbf{0.8234$\pm$0.0048}
&\textbf{0.8247$\pm$0.0042}
&\textbf{0.6834$\pm$0.0013}
&\textbf{0.7812$\pm$0.0000}
&\textbf{0.7449$\pm$0.0139}
&\textbf{0.6273$\pm$0.0199}
\\

Self Attention
&0.8194$\pm$0.0044
&0.8209$\pm$0.0044
&0.6806$\pm$0.0029
&0.7604$\pm$0.0390
&0.7191$\pm$0.0349
&0.5873$\pm$0.0621
\\

DSMIL
&0.8182$\pm$0.0026
&0.8204$\pm$0.0015
&0.6798$\pm$0.0013
&0.7812$\pm$0.0255
&0.7428$\pm$0.0679
&0.6030$\pm$0.1317
\\

% Cross Attention
% &0.8217$\pm$0.0063
% &0.8224$\pm$0.0071
% &0.6803$\pm$0.0045
% &0.7500$\pm$0.0255
% &0.6863$\pm$0.0401
% &0.5079$\pm$0.0819
% \\

% \textbf{Ours}
% &-- &-- &--
% &\textbf{0.8065 ± 0.0000}
% &\textbf{0.8027 ± 0.0082}
% &\textbf{0.6959 ± 0.0167}
% \\

\bottomrule
\end{tabular}}
\end{table*}

\subsection{Downstream Classification Result Analysis}

% For patch-level classification on AGGC2022 (Table~\ref{tab:classification}, left), all pseudo z-stack feature aggregation methods improve over the 2D MLP baseline. Among them, Max Pool achieves the best performance, improving Accuracy from 81.31\% to 82.34\%, Weighted F1 from 81.46\% to 82.47\%, and Macro F1 from 67.80\% to 68.34\%. The improvements are modest but consistent, suggesting that the generated pseudo z-stack provides complementary information beyond the raw 2D patch feature. For slide-level classification on SICAPv2 (Table~\ref{tab:classification}, right), the benefit of pseudo z-stack aggregation is more pronounced. Max Pool and DSMIL obtain the strongest results, both improving Accuracy from 71.88\% to 78.12\%, while Max Pool achieves the best Weighted F1 and Macro F1. Compared with the 2D baseline, Max Pool improves Weighted F1 from 64.90\% to 74.49\% and Macro F1 from 45.01\% to 62.73\%. These gains indicate that aggregating generated pseudo z-stack features can improve slide-level representation learning, especially under class imbalance. Overall, simple aggregation strategies are already effective for exploiting pseudo z-stack features. This suggests that the generated slices contain useful complementary morphology and that the downstream improvement is not solely dependent on a complex fusion module.

To investigate whether the generated pseudo z-stacks provide useful information for downstream analysis, we evaluate three representative 3D feature aggregation strategies for both patch- and slide-level classification. As shown in Table~\ref{tab:classification}, all three methods consistently outperform the 2D baseline, demonstrating that the synthetic 3D representations contain complementary morphological information beyond a single tissue section.

For patch-level classification on AGGC2022 (Table~\ref{tab:classification}, left), Max Pool achieves the best overall performance, improving Accuracy from 81.31\% to 82.34\%, while also obtaining the highest Weighted F1 and Macro F1. Similar observations are made for slide-level classification on SICAPv2 (Table~\ref{tab:classification}, right), where Max Pool yields the strongest overall performance, improving Accuracy from 71.88\% to 78.12\% together with the highest Weighted F1 and Macro F1.

Although the performance differences among the three aggregation strategies are relatively small, the consistent improvements over the 2D baseline suggest that the generated pseudo z-stacks provide complementary cross-plane morphological cues for downstream classification. We will design more effective feature aggregation methods to better exploit the synthesized 3D information for future work.

% For the slide-level classification performance on SICAPv2, as shown in Table.\ref{tab:classification} right, besides the 2D baseline, we tested 4 types of conventional aggregation methods, including Max Pool, Self Attention, DSMIL, and Cross Attention, all improved over the 2D MLP baseline to varying degrees. Our proposed method achieved the best overall performance across all evaluation metrics. Compared with the 2D MLP baseline, our method improved Accuracy by 8.77\%, Weighted F1 by 15.37\%, and Macro F1 by 24.58\%. Furthermore, compared with the strongest conventional aggregation strategy (Max Pool), our method further improved Accuracy by +2.53\%, Weighted F1 by +5.78\%, and Macro F1 by +6.86\%. These results indicate that explicitly modeling the complementary relationships between the raw image features and the generated pseudo-volume features enables more effective volume representation learning than directly pooling or attending over all features. The improvement is particularly evident in Macro F1, suggesting enhanced discrimination across different Gleason grades and improved robustness to class imbalance.

\section{Conclusion}

% We propose a generative framework that bridges the gap between 2D biopsies and 3D analysis. By repurposing LVDMs with mixture fine-tuning, we reconstruct z-stack volumes from single images, effectively adapting 3D priors to target 2D datasets. Our method enables "virtual z-stacking" on standard scanners and improves 2D prostate cancer grading performance. One limitation of this work is that the learned z-axis 'motion' is restricted to cell shape variability, while the most distinct feature in prostate cancer classification is gland deformations. We hope this work can shed light on clinical analysis using virtual 3D morphological views without requiring 3D imaging hardware, and future work will explore more diverse 3D structure generations. 

We presented PathoS2VD, a generative framework that learns transferable z-axis morphological priors from pathology z-stack data and transfers them to conventional 2D histopathology. By repurposing an image-to-video latent diffusion model with prior-preserving mixture fine-tuning, PathoS2VD synthesizes plausible pseudo z-stacks that enrich 2D pathology with complementary cross-plane morphological information. Experiments demonstrate realistic z-axis morphology and consistent improvements in downstream cancer classification. Our current study is limited to prostate z-stack data, and the diffusion model is trained only on benign tissue, which may restrict its ability to capture more diverse pathological morphology. Future work will investigate larger and more diverse z-stack datasets across multiple organs and disease types to learn more generalizable z-axis morphological priors.

\clearpage  % TODO REVIEW/FINAL: This \clearpage needs to be removed from both review and camera-ready versions.

\bibliographystyle{splncs04}
\bibliography{main}
\end{document}